%% file: nasgd.tex
\documentclass{article}

\usepackage[linesnumbered,ruled,vlined]{algorithm2e}
\input{header}

\SetKwInput{KwInput}{Input}
\SetKwInput{KwOutput}{Output} 

\title{Reparametrizing gradient descent}
\author{David Sprunger}
\date{\today}

\begin{document}
\maketitle

\textbf{Abstract.} In this work, we propose an optimization algorithm
which we call norm-adapted gradient descent. This algorithm is similar
to other gradient-based optimization algorithms like Adam or Adagrad in
that it adapts the learning rate of stochastic gradient descent at each
iteration. However, rather than using statistical properties of
observed gradients, norm-adapted gradient descent relies on a
first-order estimate of the effect of a standard gradient descent
update step, much like the Newton-Raphson method in many dimensions.
Our algorithm can also be compared to quasi-Newton methods, but we seek
roots rather than stationary points. Seeking roots can be justified by
the fact that for models with sufficient capacity measured by
nonnegative loss functions, roots coincide with global optima. This
work presents several experiments where we have used our algorithm; in
these results, it appears norm-adapted descent is particularly strong
in regression settings but is also capable of training classifiers.

\section{Introduction}

We propose an optimization algorithm closely related to gradient
descent. In short, to find the minimizing input of a nonnegative
real-valued function $f$, our proposal replaces the standard gradient
update step
\begin{eqnarray}
\theta_{i+1} \gets \theta_i - \eta\cdot \gradfth
\label{eq:sgd_step}
\end{eqnarray}
with a new step
\begin{eqnarray}
\theta_{i+1} \gets \theta_i - \alpha\cdot\frac{f(\theta_i)}{\norm{\gradfth}^2} \gradfth
\label{eq:nasgd_step}
\end{eqnarray}

We call the resulting algorithm ``norm-adapted gradient descent'' and
call gradient descent with the update step (\ref{eq:sgd_step})
``standard gradient descent'' for purposes of disambiguation. This
paper gives reasons for this modification, analyzes some strengths and
weaknesses of this proposal, and presents some experimental evidence
on its performance.

\begin{algorithm}[ht]
\SetAlgoLined
\DontPrintSemicolon
\KwInput{$f: \R^k \to [L, \infty)$ a differentiable function, 
$L \in \R$ a lower bound on the output of $f$ (usually $L = 0$),
$\theta_0 \in \R^k$ an initial guess,
$\alpha \in [0, 2]$ a percent decrease target}
\KwOutput{$\theta_N\in \R^k$, an input value making $f(\theta_N) \approx L$}
 $i \gets 0$\;
 \While{progress\_insufficient()}{
  $\theta_{i+1} \gets \theta_i - \min(1, \alpha \cdot \frac{f(\theta_i) - L}{\norm{\gradfth}^2}) \cdot \gradfth$
  \tcp*{update step (2)}
  $i \gets i + 1$\;
 }
 \Return $\theta_i$
 \caption{Norm-adapted gradient descent}
 \label{alg:nasgd}
\end{algorithm}

One way to describe norm-adapted gradient descent is that it performs a
variant of the Newton-Raphson root-finding algorithm. Recall in the
case $f: \R \to \R$, the Newton-Raphson method approximates $f(\theta)$ 
as the first-order function
  $f(\theta_i) + (\theta - \theta_i) \cdot f'(\theta_i)$ 
around $\theta_i$ and then chooses 
  $\theta_{i+1} \gets \theta_i - \frac{f(\theta_i)}{f'(\theta_i)}$, 
the value of $\theta$ that would make this approximation 0.

In the case $f: \R^n \to [0, \infty)$, we first restrict our search for a
root from $\theta_i$ to the $-\gradfth$ direction. The appropriate 
Newton-Raphson update is then
  $\theta_{i+1} \gets \theta_i - \frac{f(\theta_i)}{\norm{\gradfth}}
   \cdot \frac{\gradfth}{\norm{\gradfth}}$.
We introduce a hyperparameter $\alpha$, analogous to the learning rate
in standard gradient descent, after which we have the update step
(\ref{eq:nasgd_step}) above.

To see that (\ref{eq:nasgd_step}) is the appropriate Newton-Raphson
update, consider the first-order approximation of the effect of the
update step (\ref{eq:sgd_step}) on the value of $f$. From the fact that
$\frac{\partial}{\partial \eta} f(\theta_i + \eta\cdot\gradfth) =
\norm{\gradfth}^2$, 
\begin{eqnarray}
f(\theta_i - \eta\cdot\gradfth) = f(\theta_i) - \eta\cdot\norm{\gradfth}^2 + o(\eta^2).
\label{eq:fo_sgd}
\end{eqnarray}
The update step (\ref{eq:nasgd_step}) chooses the
learning rate $\eta \teq \alpha\cdot\frac{f(\theta_i)}{\norm{\gradfth}^2}$
which reduces the right hand side of the approximation
(\ref{eq:fo_sgd}) to $(1 - \alpha)f(\theta_i)$ plus the error term.

This lends a natural interpretation to norm-adapted gradient descent:
it adjusts the learning rate at every step to attempt to reduce the
value of the target function by a fixed fraction $\alpha$. This in turn
suggests $\alpha$ should be drawn from $[0, 1]$. (In practice, we have
found values of $\alpha$ up to 2 can be useful.) This is one reason for
the title of this paper---norm-adapted descent is similar to stochastic
gradient descent but with a different hyperparameter which has a
different interpretation.\\

When the gradient has a small norm, the update step
(\ref{eq:nasgd_step}) has some numerical instability, so we bound the
scalar coefficient of the gradient with a maximum value of 1, as seen
on line 3 in Algorithm~\ref{alg:nasgd}. The threshold value of 1 was
chosen since learning rates are never, to our knowledge, chosen outside
[0, 1] and this factor can be seen as the equivalent of a learning rate
in standard gradient descent.

Norm-adapted gradient descent is a root-finding algorithm. When using
it in a machine learning context we are taking advantage of the fact
that most error functions take nonnegative values. In models with
sufficient capacity, the true global minimum value will be close to
zero, and this coincidence between optimum values and roots allows us
to use root-finding algorithms. When optimizing functions which have
some known non-zero lower bound $L$, the numerator of the update in
(\ref{eq:nasgd_step}) can be modified to $f(\theta_i) - L$. Further
adaptation is required for situations where the global minimum value is
not known in advance; this is one situation where $\alpha > 1$ can be
useful.

The numerator $f(\theta_i)$ in (\ref{eq:nasgd_step}) provides a measure
of global progress to the minimum value, making the algorithm take
bigger steps when the current function value is large. Since 
norm-adapted gradient descent has access to this global information, it
is better able to distinguish local minima from global minima and
behaves rather differently from ordinary gradient descent near local
minima and saddle points.

Another important property of norm-adapted descent is its invariance to
scaling of the target function---it treats $f(\theta)$ and $k\cdot
f(\theta)$ exactly the same. Steps taken by standard gradient descent
for the latter are scaled by a factor of $k$ compared to the former,
and the optimal learning rates will differ by a factor of $k$ as well.
We have found that norm-adapted descent generally requires less
hyperparameter tuning and suspect this property may be part of the
reason why.

\section{Simple functions}

In this section, we examine the performance of these gradient descent
optimizers on some simple functions. For now, we restrict our attention
to gradient descent and our norm-adapted variant. As we progress to
more practical benchmarks, we will consider other common optimizers
like Adam.

We consider two functions: $q(x, y) \teq 8x^2 + \frac{1}{2}y^2$ and
$r(x, y) \teq (1 - x)^2 + 100(y - x^2)^2$, the Rosenbrock function.
These both present ``valley problems'' to gradient-based algorithms,
which typically make rapid progress moving down the slope of the valley
but struggle along the valley floor. The ``valleys'' for these
functions are the $y$-axis for the quadratic function and a parabolic
valley along $y = x^2$ for the Rosenbrock function. The global minimum
for each of these functions is 0, located at (0, 0) for $q$ and at $(1,
1)$ for $r$.

\subsection{Results summaries}

Table~\ref{table:q_steps} presents data about the performance of
standard gradient descent (SGD) and norm-adapted gradient descent
(NaSGD) optimizers when attempting to find points such that $q(x, y)
\leq 10^{-10}$ starting from the initial point $(1, 1)$.
We give three kinds of instances for each algorithm: first the best
hyperparameter value we could find, then results with more typical
hyperparameter values that one might try first, and finally average
results from 2000 trials with hyperparameter values uniformly sampled
from an interval.

\begin{table}[h]
\begin{tabular}{|l|rrrrr||rlrl|}
\hline
 & \multicolumn{5}{l||}{Steps to reach:} & \multicolumn{4}{l|}{Time (\& scaled to best SGD run):} \\
 & $10^{-2}$ & $10^{-4}$ & $10^{-6}$ & $10^{-8}$ & $10^{-10}$ & ms/run & & $\mu$s/step & \\\hline
\hline
SGD($\eta=0.1156$) & 23 & 39 & 56 & 74 & 92  &  18.2 & (1) & 198 & (1) \\\hline
NaSGD($\alpha=1.9$) &  4  &  6 & 10 & 13 & 15  &  4.8 & (0.27) & 322 & (1.63) \\\hline
\hline
SGD($\eta=0.0100$) & 196 & 425 & 654  & 883 & 1113  &  233.5 & (12.79) & 210 & (1.06) \\\hline
SGD($\eta=0.1000$) & 20 & 42 & 64 & 86 & 107  &  21.2 & (1.16) & 198 & (1.00) \\\hline
NaSGD($\alpha=0.7$) & 12 & 22 & 32 & 42 & 53  &  16.2 & (0.89) & 306 & (1.54) \\\hline
\hline
NaSGD($\alpha \sim [0.2, 2]$) & 13.5 & 23.35 & 33.18 & 43.03 & 52.8  & 15.9 & (0.87) & 300 & (1.51) \\\hline
\end{tabular}
\caption{Steps and time to achieve $q(x_i, y_i) \leq 10^{-10}$ from $(x_0, y_0) = (1, 1)$}
\label{table:q_steps}
\end{table}

In this scenario, though a step of norm-adapted descent takes 50-65\%
longer than a step of standard gradient descent, a large reduction in
the number of steps required leads to an overall decrease in the time
required to reach the target threshold by 10-70\%. The absolute times
reported here were obtained on a 2017 MacBook Pro, 2.3 GHz i5, with
16GB RAM.

Table~\ref{table:r_steps} gives similar data for optimizing the
Rosenbrock function $r$ starting from $(-3, -4)$. In this
table, entries marked $\uparrow$ indicate the algorithm diverged
(meaning parameter values became large).

\begin{table}[h]
\begin{tabular}{|l|rrrrr||rlrl|}
\hline
 & \multicolumn{5}{l||}{Steps to reach:} & \multicolumn{4}{l|}{Time (\& scaled to best SGD run):} \\
 & $10^{2}$ & $10^{0}$ & $10^{-2}$ & $10^{-4}$ & $10^{-6}$ & ms/run & & $\mu$s/step & \\\hline
\hline
SGD($\eta=0.0003945$) & 7 & 9 & 14 & 20 & 13633  & 3956.4 & (1) & 290 & (1) \\\hline
NaSGD($\alpha=1.6$) & 5 & 10 & 57 & 166 & 236  & 84.4 & (0.02) & 358 & (1.23) \\\hline
\hline
SGD($\eta=0.0001$) & 66 & 254 & 31683 & 85861 & 142986  & 37806.6 & (9.56) & 264 & (0.91) \\\hline
SGD($\eta=0.0010$) & $\uparrow$ & $\uparrow$ & $\uparrow$ & $\uparrow$ & $\uparrow$
 & --- & (---) & --- & (---) \\\hline
NaSGD($\alpha=0.7$) & 9 & 16 & 147 & 354 & 497  & 185.9 & (0.05) & 374 & (1.29) \\\hline
\hline
NaSGD($\alpha \sim [0.2, 2]$) & 8.37 & 14.89 & 152.2 & 390.31 & 641.93  & 239.7 & (0.06) & 373 & (1.29) \\\hline
\end{tabular}
\caption{Steps and time to achieve $r(x_i, y_i) \leq 10^{-6}$
 from $(x_0, y_0) = (-3, -4)$}
\label{table:r_steps}
\end{table}

The results here are broadly similar to the quadratic function:
individual steps of norm-adapted descent are more costly than those of
standard gradient descent, but the overall time to reach the lower
values is considerably reduced since fewer steps are required. The
per-step overhead involved in norm-adapted descent is the cost to
compute the squared norm of the gradient, since all the other
quantities are computed in standard gradient descent. \\

Hyperparameter tuning for norm-adapted descent is relatively easy:
choosing a random $\alpha \in [0.2, 2]$ yields, on average, performance
competitive with the best version of standard gradient descent we could
find \textbf{in both scenarios}. In contrast, learning rates for
gradient descent are sensitive to the scale of the function, and no
learning rate works out-of-the-box in both cases. Learning rates in the
$10^{-3}$ or $10^{-4}$ range are intolerably slow for the quadratic
function, while learning rates in the $10^{-1}$ or $10^{-2}$ range for
the Rosenbrock function cause immediate divergence. The Rosenbrock
function, a common test function in optimization, is particularly
tricky: gradient descent seems to start diverging with learning rates
higher than $0.0003948$, while our best-performing learning rate is
just barely under this value.

\subsection{Vector fields}

\begin{wrapfigure}{r}{0.45\textwidth}
\vspace{-1.2cm}
\includegraphics[scale=0.30]{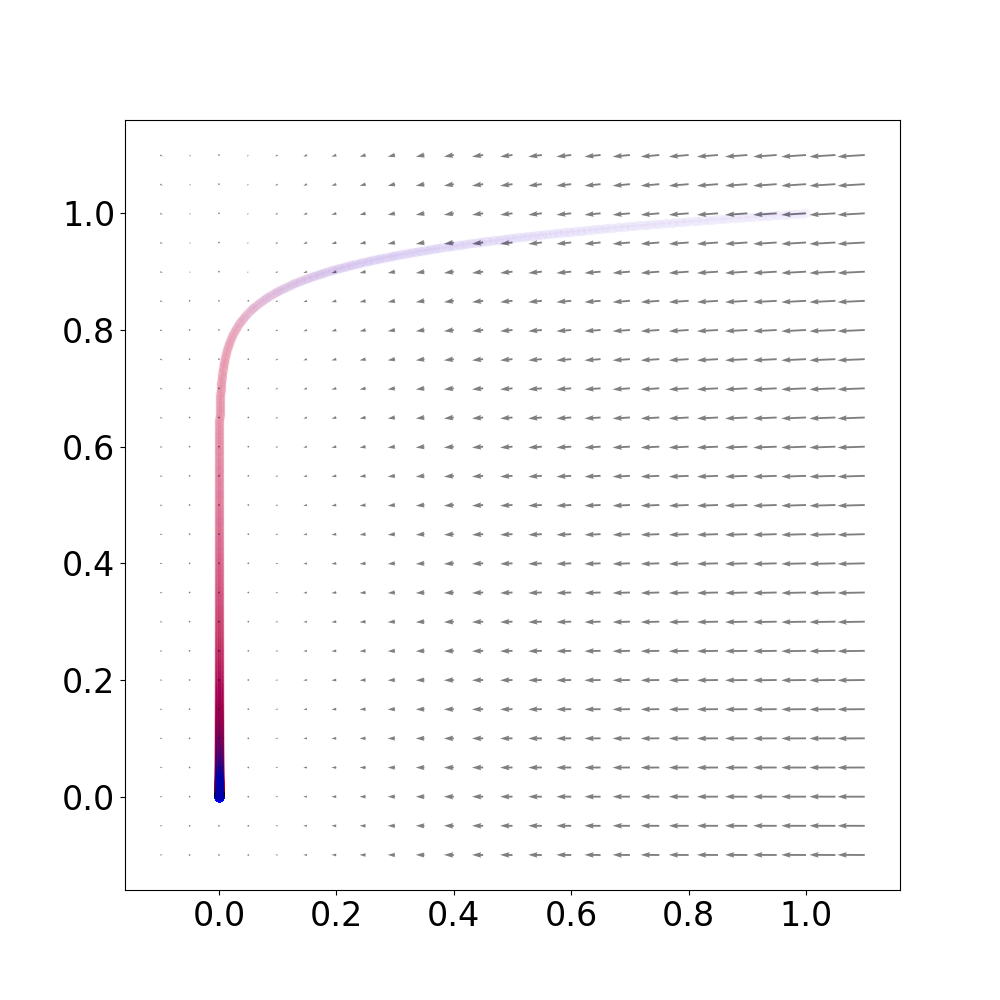}
\vspace{-0.5cm}
\caption{Comparing limit behavior}
\label{fig:limit_beh}
\end{wrapfigure}

We can get some insight into why these two algorithms perform
the way they do by considering what they do in the limit. In
figure~\ref{fig:limit_beh}, we have plotted the vector field $\vec{x}
\mapsto -\nabla q(\vec{x})$, together with two series of very light
circles. In light blue circles are the first 6000 values NaSGD($\alpha
= 0.007$) goes through while optimizing $q$ from $(1, 1)$. In light red
are the first 6000 values SGD($\eta = 0.001$) visits while optimizing
the same function from the same point.

The fact that there is a single curve from $(1, 1)$ to $(0, 0)$
indicates that in the limit (as $\alpha, \eta \to 0$), these two
algorithms transit through the same points. The fact that this curve is
not a uniform color also contains important information. Where the
curve is darker, at least one of the algorithms is spending more time:
norm-adapted descent is spending more time in the bluer regions, and
standard gradient descent spends more time in redder regions. 

\begin{figure}[h!]
\vspace{-0.3cm}
\includegraphics[scale=0.30]{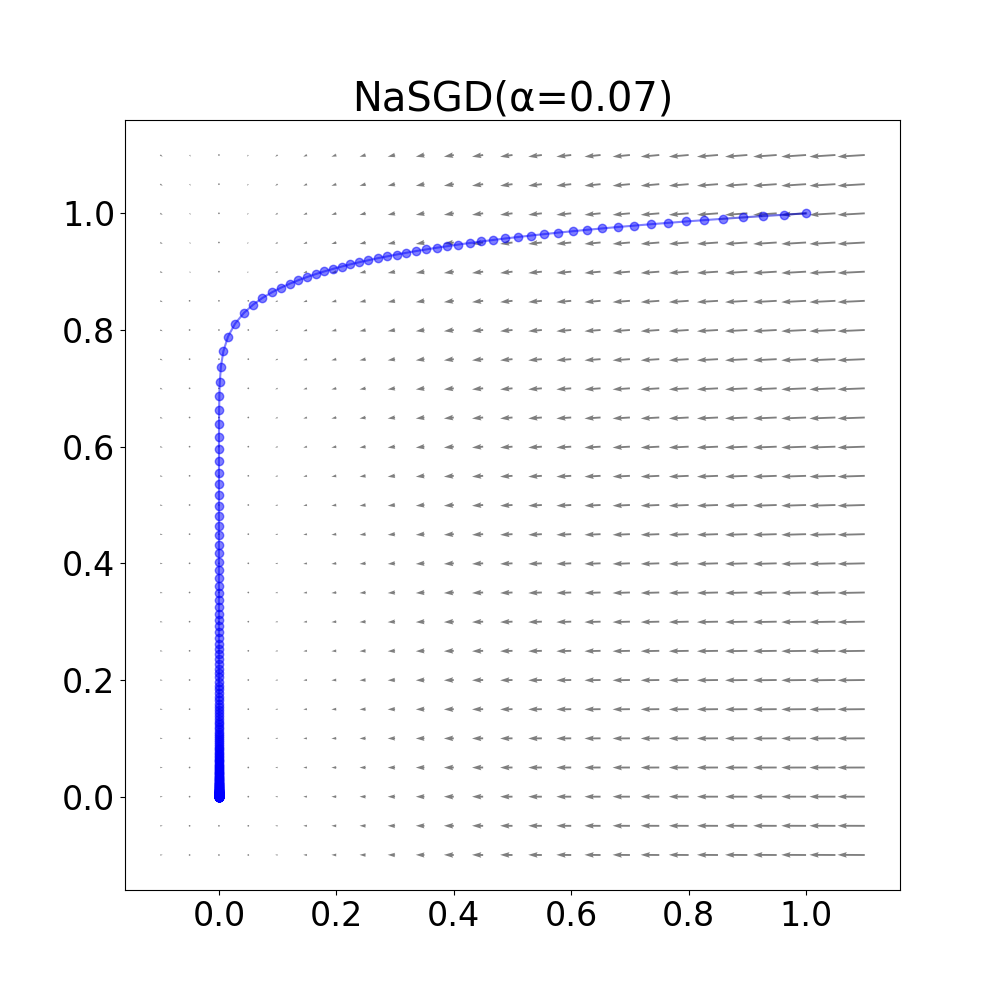}
\includegraphics[scale=0.30]{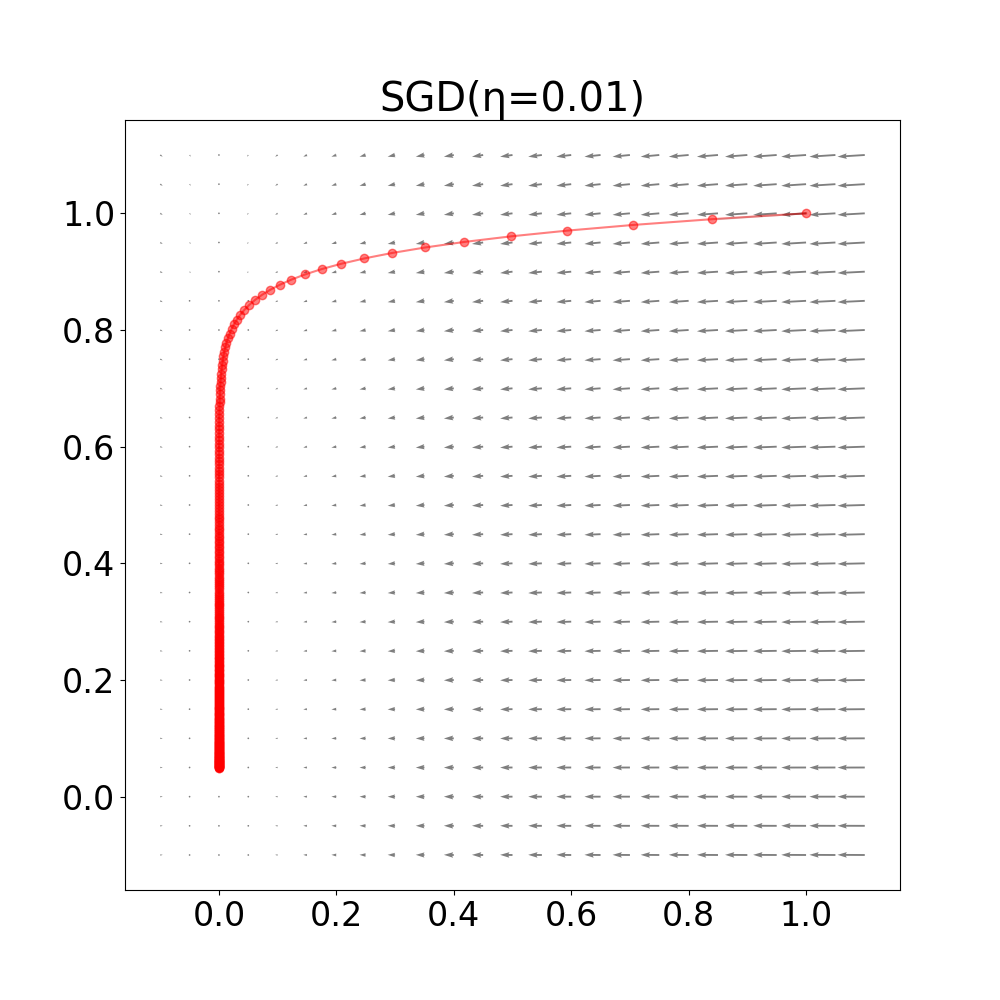}
\vspace{-0.5cm}
\caption{Points visited by norm-adapted descent and gradient descent}
\label{fig:limit_discrete}
\end{figure}

We can illustrate this further by plotting the points each algorithm
visits when these hyperparameters are multiplied by 10
(Figure~\ref{fig:limit_discrete}). As we can see, norm-adapted descent
takes much smaller steps initially, taking much longer to reach an
$x$-value under 0.2. However, as the two processes begin moving mostly
in the $y$-direction, norm-adapted descent picks up speed: individual
points are distinguishable whereas in standard gradient descent they
are not.

There is a nice reason for these phenomena. Both types of gradient
descent are finding approximations to integral flows for two vector
fields which have the same \textbf{direction} at every point ($\vec{x}
\mapsto -\nabla q(\vec{x})$ for standard gradient descent and $\vec{x}
\mapsto - \frac{q(\vec{x})}{\norm{q(\vec{x})}^2}\nabla q(\vec{x})$ for
norm-adapted descent), but which may have different \textbf{lengths}.
The fact that these vector fields have the same direction everywhere
means their integral flows pass through the same set of points, and the
fact that these fields usually have different lengths mean the two
integral flows pass through those points at different speeds. This is
the second justification for our title---the two versions of gradient
descent effectively transit the same curve, but with different
parameterizations in time.

In particular, since standard gradient descent follows the gradient
vector field, it moves quickly through regions where the gradient has a
large norm and slowly through regions where the norm is small. This is
why standard gradient descent makes more rapid progress in the
$x$-direction while optimizing $q$---the gradient of $q$ along the
$y$-axis is comparatively small.
Norm-adapted descent behaves differently. The length of its vector
field at $\vec{x}$ is $\frac{q(\vec{x})}{\norm{q(\vec{x}))}}$, so
larger gradients \textbf{slow down} norm-adapted descent and small
gradients accelerate it. This is the reason it is able to transit the
``valley floor'' along the $y$-axis quickly.

\begin{wrapfigure}{l}{0.45\textwidth}
\vspace{-0.5cm}
\includegraphics[scale=0.30]{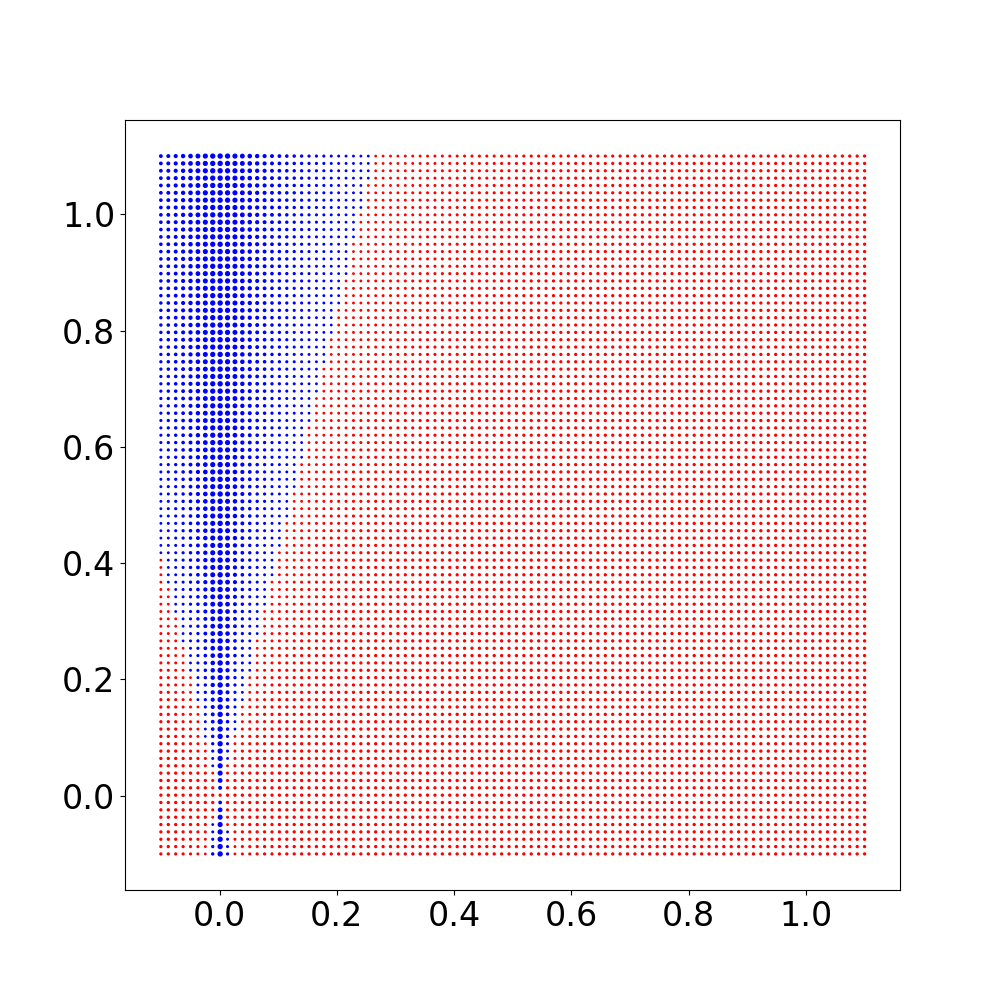}
\vspace{-0.8cm}
\caption{Speed advantages}
\label{fig:speed}
\vspace{-0.3cm}
\end{wrapfigure}

The net effect of all of this is that norm-adapted descent moves more
slowly in regions where standard gradient descent moves quickly and
conversely. While this might seem to be a balanced trade, it is
decidedly to norm-adapted descent's advantage: decreasing the
difference between the speeds in the two regions considerably raises
their harmonic mean.

Figure \ref{fig:speed} depicts the ratio between the lengths of the two
vector fields at each point in space. We plot a red circle at points
where standard gradient descent is following a longer vector and
conversely a blue circle where norm-adapted descent is longer. The area
of the circle is proportional to the ratio of the larger to the
smaller. We also include a factor of the ratio of the optimal learning
rates, since the magnitudes of the learning rates affect the speed of
the convergence.

Standard gradient descent has a consistent advantage away from the
$y$-axis, following a gradient field which is moving about twice as
fast. Closer to the $y$-axis, norm-adapted descent acquires a large
advantage, following a gradient field which is about eight times larger
than standard gradient descent.

\section{Dense layer matching task}

In this section, we continue our examination of norm-adapted gradient
descent in increasingly applied contexts. Here we will look at a
somewhat contrived machine learning task: a layer recovery.  In this
task, we randomly initialized a dense layer $\R^{10}\to\R^4$ with a
$\tanh$ activation function. We then randomly sampled 200 input points
from $[0, 1]^{10}$ and ran the dense layer on these points to get
corresponding outputs. We split this artificial dataset into two
equally sized subsets, one for training and one for testing. 

We randomly initialized layers of the same shape and trained them on
the test set in an attempt to recover the original layer. We measured
progress on this task as the average distance between the expected and
predicted values on the test set. We used various optimizers to train
each network for 45 passes through the data with minibatches of size 1.
Therefore, in the graphs below, 100 training steps correspond to an
epoch of training. We repeated this experiment 50 times and report on
the averages achieved in these 50 runs. \\

This scenario is unusually well-suited for neural networks; it is much
more typical to have no guarantees about how well the data-generating
process under consideration can modeled by our network architecture.
Here, we know that our model has the exact right amount of capacity and
is structured in the best possible way. The primary challenge here is
using a relatively limited amount of data to recover the parameters.
However, common optimization algorithms struggle to achieve average
distances lower than $10^{-3}$ between the actual and expected points.

\subsection{Results summaries}

\begin{figure}[h!]
\begin{center}
\includegraphics[scale=0.3]{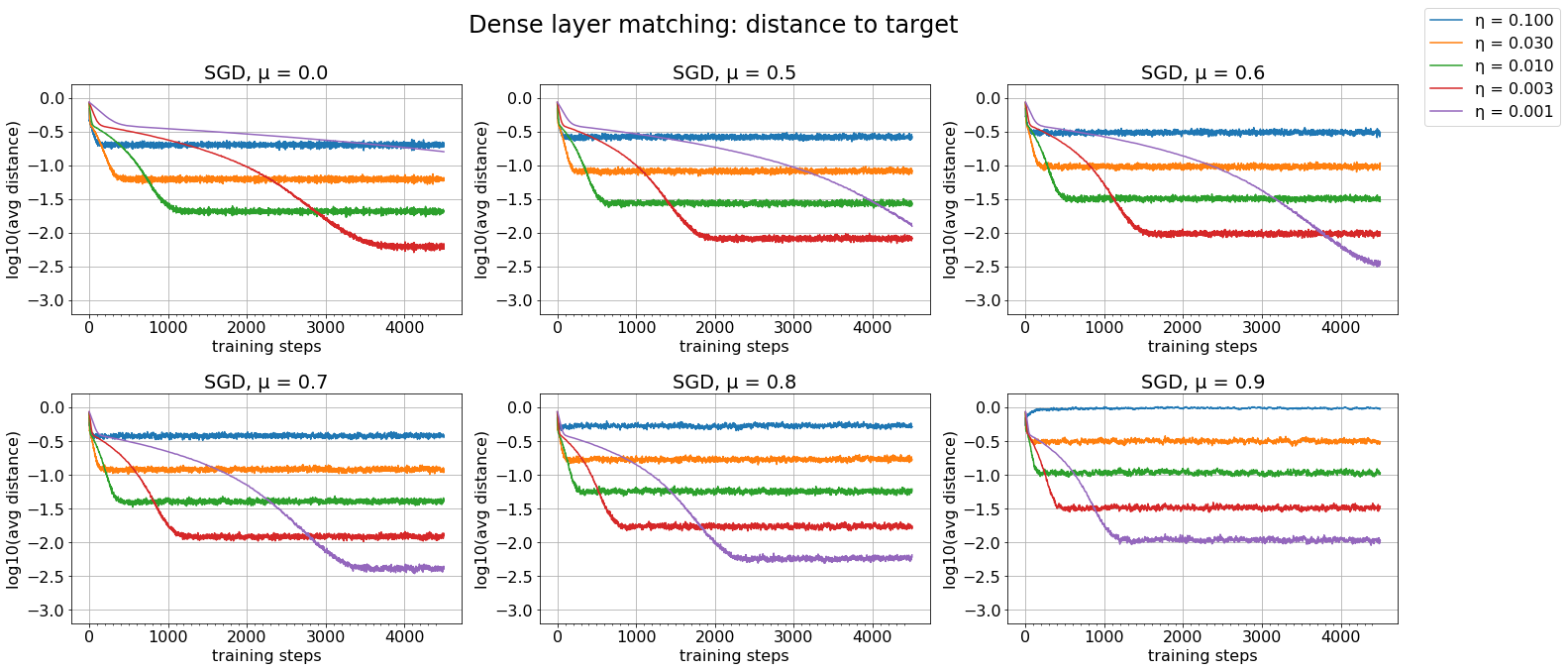}
\end{center}
\vspace{-0.5cm}
\caption{Standard gradient descent with momentum in the dense layer matching task}
\label{fig:copy_sgd}
\end{figure}

In figure \ref{fig:copy_sgd}, we plot the base-10 logarithm of the
average distance between actual test values and the values produced by
the networks trained by standard gradient descent with various learning
rates as a function of the number of training steps. Each plot
corresponds to a different value of momentum, indicated in the title of
the plot as $\mu$; each series corresponds to a different learning
rate, indicated in the legend as $\eta$.

As we can see, each learning rate seems to have a saturation point
beyond which further training does not improve performance. Higher
learning rates reach their saturation points quicker, while lower
learning rates saturate at better performance levels. Raising the value
the momentum parameter causes overall gains in speeds of convergence
but carries some cost in terms of final performance, findings
consistent with \cite{SutskeverMomentum}. 

\begin{figure}[h!]
\begin{center}
\includegraphics[scale=0.28]{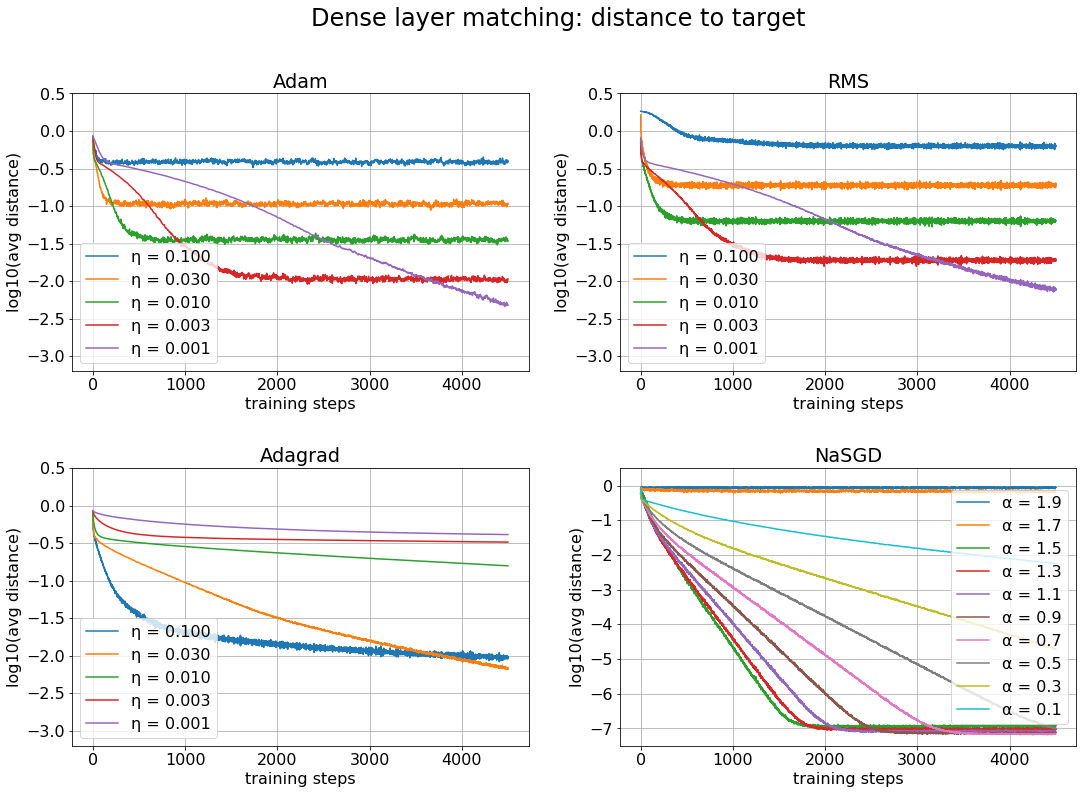}
\end{center}
\vspace{-0.5cm}
\caption{Adam, RMSprop, Adagrad, and NaSGD in the dense layer matching task}
\label{fig:copy_other}
\end{figure}

Other common optimization algorithms (Adam, RMSprop, and Adagrad)
similarly struggle to break $10^{-3}$, as shown in figure
\ref{fig:copy_other}. Norm-adapted descent achieves average distances
lower than $10^{-3}$ with progress stabilizing around $10^{-7}$ for
$\alpha \in [0.5, 1.5]$. Norm-adapted descent with $\alpha \in [0.1,
0.3]$ is competitive with standard algorithms, and $\alpha \geq 1.7$
appears not to make progress. 

\subsection{Translating between hyperparameters}
\label{ssec:rosetta}

A useful coincidence between standard gradient descent and our
norm-adapted variant is the fact that they follow the same direction
field. This means that for each step taken by standard gradient 
descent we can calculate the equivalent $\alpha$ which would produce
the exact same step when used in norm-adapted descent, and conversely
we can calculate the equivalent $\eta$ for each step taken by
norm-adapted descent. We present the average such equivalent parameters
over the 50 runs in the first two plots in figure \ref{fig:rosetta}
below. 

\begin{figure}[h!]
\begin{center}
\includegraphics[scale=0.33]{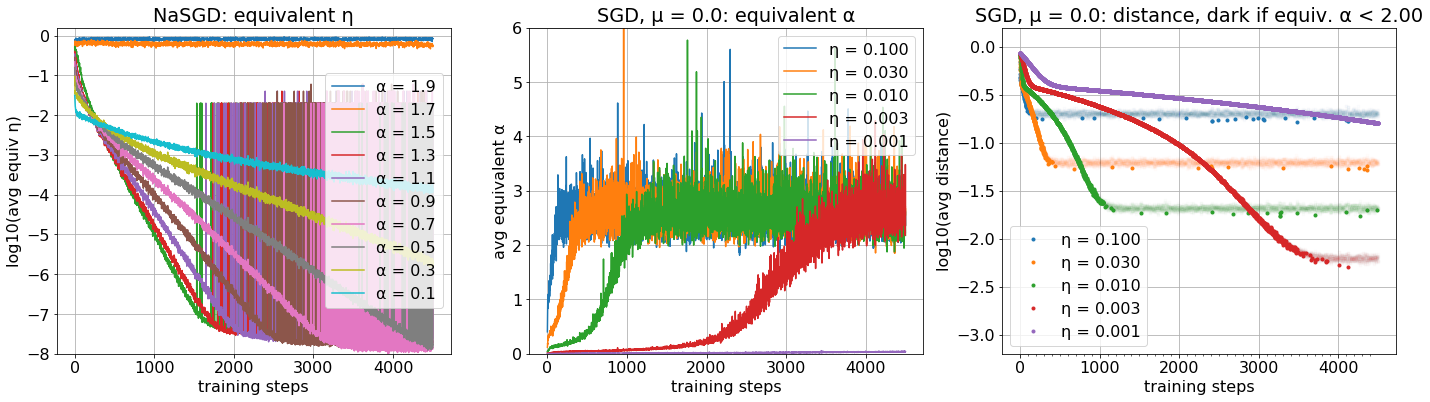}
\end{center}
\vspace{-0.5cm}
\caption{Rosetta stone: equivalent hyperparameters for other algorithm}
\label{fig:rosetta}
\end{figure}

As we see in the first plot, though norm-adapted descent could end up
having equivalent learning rates anywhere in the $[0, 1]$ interval at
any time, it generally decreases the learning rate from 1 to $10^{-8}$
as its performance improves. One way to interpret this is that it is
performing an adaptive learning rate scheduling based on the current
loss value and gradient.

The second plot shows standard gradient descent starts with low
equivalent $\alpha$, which tends to increase during the course of
training, settling above 2. This means that the steps taken by standard
gradient descent should result in a 200-300\% decrease in loss
according to the estimate (\ref{eq:fo_sgd}). This is obviously
overambitious; loss can decrease by at most 100\% for nonnegative
functions.

Though we have observed $\alpha \in [0, 2]$ can be performant for
norm-adapted descent, rarely has $\alpha > 2$ resulted in good
performance. Thus, this ``equivalent $\alpha$'' statistic may be a good
indicator for a high learning rate in gradient descent. In the third
plot of figure \ref{fig:rosetta}, we show the same accuracy series as
in the first plot in figure \ref{fig:copy_sgd}, but fade the points if
the equivalent $\alpha$ in that step is greater than 2. The result
suggests a general rule: gradient descent makes progress in this task
when the equivalent $\alpha$ value is below 2 and stops making progress
when its equivalent $\alpha$ value is above 2.

The diagnostic role of this translation works in the other direction
too: norm-adapted descent makes steady progress while its equivalent
learning rate decreases steadily. After converging, the equivalent
learning rate seems to oscillate in magnitude quite significantly. This
information could possibly used to design a stopping tactic for
norm-adapted descent.

\subsection{The utility of the Newton-Raphson estimate}
\label{ssec:hybrids}

There is a natural question at this point about the fundamental reason
for the performance differences between gradient descent and the
norm-adapted variant. The equivalent $\eta$ graph in figure
\ref{fig:rosetta} suggests norm-adapted descent has an exponentially
decreasing learning rate, and scheduling a learning rate to follow such
a function is a known technique which often improves performance at the
cost of tuning further hyperparameters (namely the base of the
exponential decay). So, is adapting the learning rate using the
Newton-Raphson estimate actually important to the performance of
norm-adapted descent, or is it basically a mimicking mechanism for
learning rate scheduling?

To address this question, we consider two different kinds of
optimizers. In the first class, we fit an exponential to the average
equivalent learning rate of norm-adapted descent ($\alpha = 0.7$) in
this task and use this exponential as the learning rate schedule for an
otherwise standard gradient descent. In the second class, we do a
standard gradient descent while monitoring its equivalent $\alpha$
value. If that $\alpha$ value exceeds some threshhold too often, we
decrease the learning rate. If norm-adapted descent gets its
performance by  somehow choosing a good schedule, we would expect the
first class to behave similarly on average to NaSGD($\alpha=0.7$). If
the Newton-Raphson estimate provides a useful signal for reducing
learning rate adaptively, we would expect the second class to do well.

In the end, the first class, imitating norm-adapted descent's average
equivalent learning rate, does not perform particularly well in this task. We
tried fitting three different exponentials---one for the best fit in the first
1500 (series labeled ``decay028''), first 600 (``decay052'') and first 200
steps (``decay090'') each. However, none performed particularly well. 

\begin{figure}[h!]
\begin{center}
\includegraphics[scale=0.33]{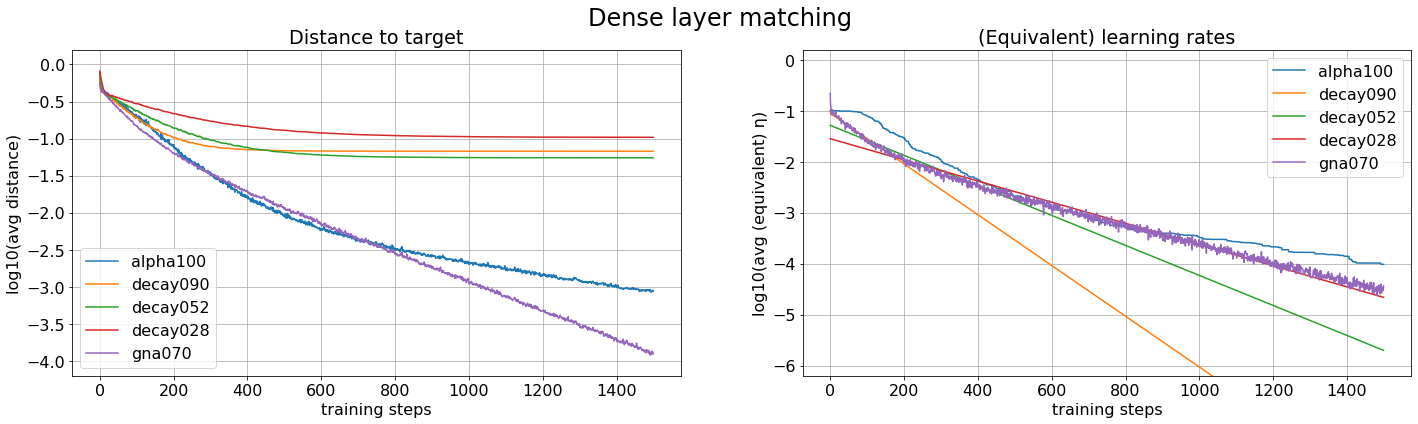}
\end{center}
\vspace{-0.5cm}
\caption{Exponential decay scheduling vs equivalent $\alpha$ monitoring}
\label{fig:copy_explain}
\end{figure}

We only tried one member of the second class (``alpha100''), where we
multiplied the learning rate (starting at $\eta = 0.1$) by
$\frac{1}{3}$ each time $\alpha \geq 2.0$ happened for 3 consecutive
steps. This simplistic policy has somewhat comparable performance to
NaSGD($\alpha = 0.7$) (``gna070''), and achieves an average distance of
$10^{-3}$ from the expected values, which is better than the standard
optimizers described above. In the plots of figure
\ref{fig:copy_explain}, we show the average accuracies and average
learning rates (or equivalent learning rates) of these five algorithms
in the first 1500 steps (15 epochs) of the same 50 runs as used in the
experiments above. 

As a result, we believe that the Newton-Raphson modeling of the
function and norm-adapted descent's continuous monitoring of this
quantity plays a critical role in its enhanced performance.

\section{MNIST}

Finally, we consider a standard problem in machine learning:
classification on the MNIST dataset \cite{mnist}. We train simple
convolutional neural networks to classify these images, with two
convolutional layers (5x5 filters with 10 then 20 features) with 2x2
max pooling layers after each convolution and ending with two dense
layers with 50 then 10 output neurons. For regularization, we included
dropout on the second convolution and first dense layers and normalized
the pixel value distribution. This is a relatively small convolutional
network, but with enough capacity to regularly classify 97\% of MNIST
images correctly.

\begin{figure}[h!]
\begin{center}
\includegraphics[scale=0.3]{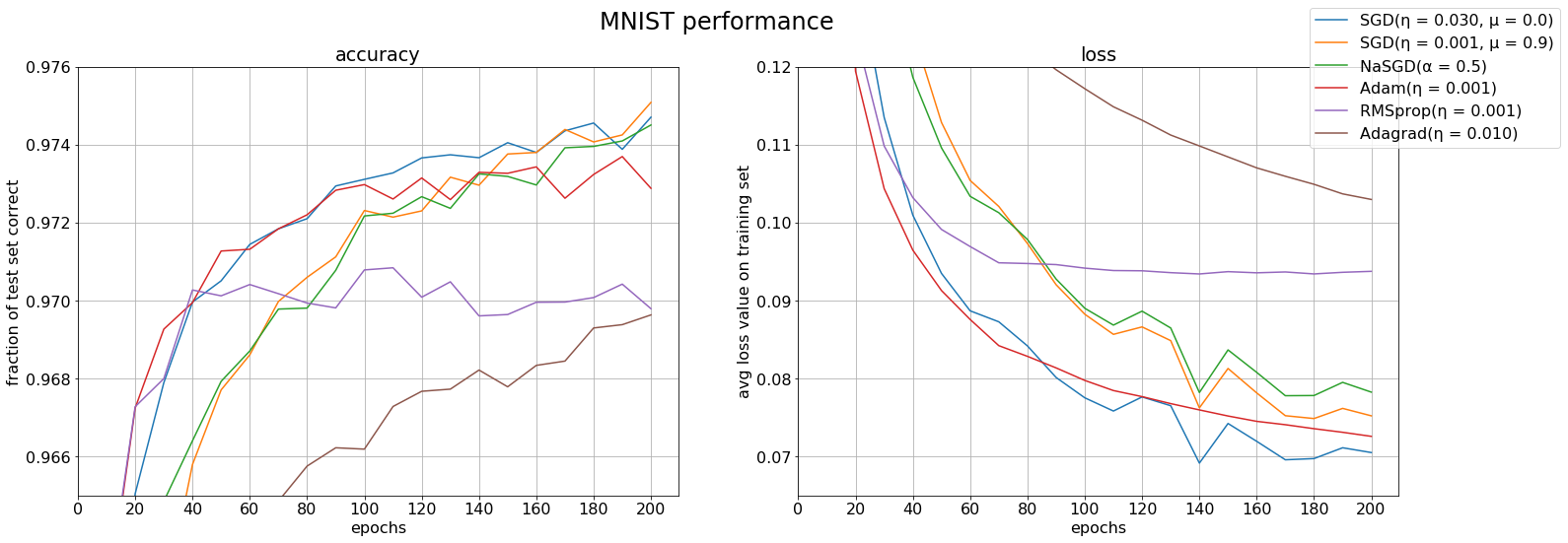}
\end{center}
\vspace{-0.5cm}
\caption{MNIST: performance metrics plotted}
\label{fig:mnist}
\end{figure}

A one-hot encoding of the correct class is used as the target value in
$\R^{10}$, and we use a cross-entropy loss function. We train the
network in minibatches of size 60 so an epoch consists of 1000
minibatches. We completed 20 different runs consisting of 200 epochs
with many different optimizers and learning rates. The data presented
in table~\ref{tab:mnist} and figure~\ref{fig:mnist} are averages of
these 20 runs for selected high-performance learning rates for each
optimizer.

Our findings largely fit with the results of Kingma and
Ba~\cite{KingmaAdam}, though our model is less complex. Adam shows
clear performance benefits over RMSprop and Adagrad in both accuracy
and loss. They consider SGD with Nesterov momentum, while we present
SGD with ordinary momentum, so those series are less comparable. While
SGD performs best in this task, norm-adapted gradient descent is
certainly competitive with SGD and Adam and achieves an edge over
RMSprop and Adagrad.

\begin{table}[h]
\begin{tabular}{llcll}
Optimizer name & Accuracy (higher better) & & Optimizer name & Loss (lower better) \\ \hline
SGD($\eta$ = 0.001, $\mu$ = 0.9) & 0.9751 & & SGD($\eta$ = 0.03, $\mu$ = 0.0) & 0.0705 \\
SGD($\eta$ = 0.03, $\mu$ = 0.0) & 0.9747  & & Adam($\eta$ = 0.001) &  0.0726 \\
NaSGD($\alpha$ = 0.5) & 0.9745            & & SGD($\eta$ = 0.001, $\mu$ = 0.9) & 0.0752 \\
Adam($\eta$ = 0.001) & 0.9729             & & NaSGD($\alpha$ = 0.5) & 0.0783 \\
RMSprop($\eta$ = 0.001) & 0.9698          & & RMSprop($\eta$ = 0.001) & 0.0938 \\
Adagrad($\eta$ = 0.01) & 0.9696           & & Adagrad($\eta$ = 0.01) & 0.103
\end{tabular}
\caption{MNIST: selected optimizers sorted by two performance metrics after 200 epochs of training}
\label{tab:mnist}
\end{table}

\section{Related work}

Norm-adapted gradient descent is a gradient-based optimization
algorithm used for training neural networks and is naturally related to
other works which propose algorithms in this same class. Adam
\cite{KingmaAdam}, Adagrad \cite{DuchiAdagrad}, RMSprop
\cite{TielemanRMSprop}, Adadelta \cite{ZeilerAdadelta}, gradient
descent with momentum and Nesterov updates
\cite{dozat2016incorporating,SutskeverMomentum} are some examples of
these kinds of algorithms, commonly used and available in many deep
learning frameworks.

Each of these optimization algorithms modify SGD using statistical
properties of the gradients seen in the training process, such as
weighted means of past gradients (SGD with momentum), raw second
moments (Adagrad, RMSprop, and Adadelta), or both (Adam). Norm-adapted
descent relies on the estimate of an update's effect (\ref{eq:fo_sgd})
and uses no such historical data, making it a largely orthogonal
development. This presents exciting opportunities for further research:
what is a sensible Newton-Raphson update step for these other
algorithms and how does it perform?

Another related family of algorithms are quasi-Newton methods, such as
BFGS. A common use of the Newton-Raphson method for root-finding is to
search for roots of the derivative of a function---such points include
all optima of differentiable functions. The update scheme for this
version of Newton's method is $\theta_{i+1} \gets \theta_i -
[H_f(\theta_i)]^{-1}\gradfth$, where the Jacobian is the function being
zeroed so the Hessian matrix takes the role of derivative. Computing
Hessians for high-dimensional functions is usually considered
prohibitively expensive, so quasi-Newton methods maintain
approximations to the Hessian.

Norm-adapted descent relies on a coincidence between optima and roots
which  allows us to do root-finding on the original function rather
than its derivative. This saves some of the higher-order complications
present in  quasi-Newton methods at the expense of generality, namely
the risk of failing to converge if the minimum value of the function is
not known in advance. The nonnegativity of most loss functions in
machine learning helps mitigate this weakness of NaSGD.

One advantage of norm-adapted descent over these other gradient-based
algorithms, including quasi-Newton methods, is a resistance to saddle
points and local minima. Incorporating the current value of the loss
function allows norm-adapted descent to differentiate between global
minima and stationary points: areas around both have small gradient
vectors but only around the latter will also have greater loss values.
This advantage comes with an obvious drawback: if the global minimum is
not known in advance or is inaccurately approximated (e.g.~when trying
to fit a model with insufficent capacity), norm-adapted descent does
not readily settle into a global minimum.

Norm-adapted descent can also be seen as a gradient-based algorithm
which adjusts its learning rate at every step. Other works which adjust
hyperparameters in the course of training include
\cite{SchaulPesky,hyperoptimizers}. The key difference between our work
and these approaches is that our learning rate adjustment is made
based on the Newton-Raphson estimate, rather than local curvature
information or the gradient of a learning step with respect to a
hyperparameter.

\section{Conclusion and future directions}

In this work, we have described a new algorithm for training neural
networks, norm-adapted gradient descent, which is based on the
Newton-Raphson method in many dimensions. We described several
experiements demonstrating the practical performance of our algorithm.
Norm-adapted gradient descent performs particularly well in regression
tasks, posting order-of-magnitude improvements in performance over
well-known optimizers. Norm-adapted descent also performed well in the
MNIST classification task, certainly competitive with other well-known
optimizers.

As noted above, norm-adapted descent breaks from the tradition of 
using statistical information about gradients observed in the training
process and instead relies on the Newton-Raphson method to adapt the
learning rate. The orthogonality of this development leads to some
interesting questions: could similar adaptations be developed for
SGD with momentum, Adam, etc?

We also noted the translatability between the learning rates in
norm-adapted descent and stochastic gradient descent leads to some
insights about the performance of both of these algorithms. We are
interested in further exploiting these observations to develop hybrid
optimization algorithms like those alluded to in
section~\ref{ssec:hybrids}.

Finally, though norm-adapted descent shows great promise based on the
experiments we were able to conduct, it will certainly be interesting
to see its performance on more advanced tasks.

\bibliographystyle{plain}
\bibliography{nasgd}

\end{document}

%% file: header.tex
\usepackage[utf8]{inputenc}
\usepackage{latexsym}
\usepackage{amssymb}
\usepackage{amsmath}
\usepackage{xparse}
\usepackage{enumerate}
\usepackage{amsthm}
\usepackage{amsfonts}
\usepackage{multicol}
\usepackage{fullpage}
\usepackage{proof}
\usepackage{wrapfig}
\usepackage{graphicx}
\usepackage{subcaption}
\usepackage{titlesec}
\usepackage{hyperref}

\usepackage{pgf}
\usepackage{tikz}
\usetikzlibrary{automata,positioning,arrows}
\usepackage{tikz-cd}

\usetikzlibrary{shapes.gates.ee,shapes.gates.ee.IEC,shapes.geometric,shapes.gates.logic.US,shapes.gates.logic.IEC}
\usetikzlibrary{shapes.multipart}

\newcommand{\norm}[1]{\left\lVert #1 \right\rVert}

\newcommand{\R}{\mathbb{R}}

\newcommand{\teq}{\mathbin{\triangleq}}

\NewDocumentCommand{\valtostate}{m O{} O{}}{\,^{#2}\lfloor#1\rceil_{#3}}

\newcommand{\rmid}[2]{\ar@{}[#1]|-{#2}}

\newcommand{\adj}[3]{
  \ar@<+.3pc>@{->}[#1]^-{#2} 
  \ar@<-.3pc>@{<-}[#1]_-{#3} 
  \ar@{}[#1]|-{\dashv} 
}

\def\gradfth{\nabla_{\theta} f({\theta_i})}





%% file: nasgd.bbl
\begin{thebibliography}{10}

\bibitem{DBLP:conf/icml/2013}
{\em Proceedings of the 30th International Conference on Machine Learning,
  {ICML} 2013, Atlanta, GA, USA, 16-21 June 2013}, volume~28 of {\em {JMLR}
  Workshop and Conference Proceedings}. JMLR.org, 2013.

\bibitem{hyperoptimizers}
Kartik Chandra, Erik Meijer, Samantha Andow, Emilio Arroyo-Fang, Irene Dea,
  Johann George, Melissa Grueter, Basil Hosmer, Steffi Stumpos, Alanna Tempest,
  and et~al.
\newblock Gradient descent: The ultimate optimizer.
\newblock {\em arXiv:1909.13371 [cs, stat]}, Sep 2019.
\newblock arXiv: 1909.13371.

\bibitem{dozat2016incorporating}
Timothy Dozat.
\newblock {Incorporating Nesterov momentum into Adam}.
\newblock 2016.

\bibitem{DuchiAdagrad}
John~C. Duchi, Elad Hazan, and Yoram Singer.
\newblock Adaptive subgradient methods for online learning and stochastic
  optimization.
\newblock {\em J. Mach. Learn. Res.}, 12:2121--2159, 2011.

\bibitem{KingmaAdam}
Diederik~P. Kingma and Jimmy Ba.
\newblock Adam: {A} method for stochastic optimization.
\newblock In Yoshua Bengio and Yann LeCun, editors, {\em 3rd International
  Conference on Learning Representations, {ICLR} 2015, San Diego, CA, USA, May
  7-9, 2015, Conference Track Proceedings}, 2015.

\bibitem{mnist}
Yann LeCun and Corinna Cortes.
\newblock {MNIST} handwritten digit database.
\newblock 2010.

\bibitem{SchaulPesky}
Tom Schaul, Sixin Zhang, and Yann LeCun.
\newblock No more pesky learning rates.
\newblock In {\em Proceedings of the 30th International Conference on Machine
  Learning, {ICML} 2013, Atlanta, GA, USA, 16-21 June 2013\/}
  \cite{DBLP:conf/icml/2013}, pages 343--351.

\bibitem{SutskeverMomentum}
Ilya Sutskever, James Martens, George~E. Dahl, and Geoffrey~E. Hinton.
\newblock On the importance of initialization and momentum in deep learning.
\newblock In {\em Proceedings of the 30th International Conference on Machine
  Learning, {ICML} 2013, Atlanta, GA, USA, 16-21 June 2013\/}
  \cite{DBLP:conf/icml/2013}, pages 1139--1147.

\bibitem{TielemanRMSprop}
T.~Tieleman and G.~Hinton.
\newblock {Lecture 6.5---RmsProp: Divide the gradient by a running average of
  its recent magnitude}.
\newblock COURSERA: Neural Networks for Machine Learning, 2012.

\bibitem{ZeilerAdadelta}
Matthew~D. Zeiler.
\newblock {ADADELTA:} an adaptive learning rate method.
\newblock {\em CoRR}, abs/1212.5701, 2012.

\end{thebibliography}
